\documentclass[letterpaper, 10 pt, conference]{ieeeconf}  
\usepackage{graphicx,multirow}
\usepackage{booktabs}

\usepackage[dvipsnames]{xcolor}
\usepackage{algorithm}
\usepackage{algpseudocode}
\usepackage{makecell}
\usepackage{dsfont}
\usepackage{subcaption}

\IEEEoverridecommandlockouts                              

\title{\LARGE \bf
Fed-EC: Bandwidth-Efficient Clustering-Based Federated Learning For Autonomous Visual Robot Navigation
}

\author{Shreya Gummadi$^{1}$, Mateus V. Gasparino$^{1}$, Deepak Vasisht$^{2}$, and Girish Chowdhary$^{1}$
\thanks{$^{1}$Field Robotics Engineering and Science Hub (FRESH), Illinois Autonomous Farm, University of Illinois at Urbana-Champaign (UIUC), IL, USA (Email: {gummadi4@illinois.edu})}%
\thanks{$^{2}$Dept. of Computer Science, UIUC, IL, USA}%
}

\begin{document}
\maketitle

\thispagestyle{empty}
\pagestyle{empty}

\begin{abstract}

Centralized learning requires data to be aggregated at a central server, which poses significant challenges in terms of data privacy and bandwidth consumption. Federated learning presents a compelling alternative, however, vanilla federated learning methods deployed in robotics aim to learn a single global model across robots that works ideally for all. But in practice one model may not be well suited for robots deployed in various environments.  This paper proposes Federated-EmbedCluster (Fed-EC), a clustering-based federated learning framework that is deployed with vision based autonomous robot navigation in diverse outdoor environments. The framework addresses the key federated learning challenge of deteriorating model performance of a single global model due to the presence of non-IID data across real-world robots. Extensive real-world experiments validate that Fed-EC reduces the communication size by 23x for each robot while matching the performance of centralized learning for goal-oriented navigation and outperforms local learning. Fed-EC can transfer previously learnt models to new robots that join the cluster.

\end{abstract}

\section{Introduction}
Poor availability of high-speed internet is limiting outdoor robots from realizing their full potential. In today's world, robots are seamlessly deployed in diverse conditions all over the world, from bustling urban landscapes to rugged terrains in the wild. Many of these robots are using visually guided autonomy architectures powered by machine learning and self-supervision. Recent works \cite{Gasparino2022WayFASTNW}, \cite{gasparino2024wayfaster}, \cite{kouris2018learning}, \cite{Kahn2020BADGRAA}, \cite{shah2022viking} have shown that with access to large amounts of data, robots can achieve state-of-the-art navigation performance and can be deployed in various scenarios with minimum human intervention needed. These and other such methods are driving tremendous progress in self-driving cars \cite{8595590}, \cite{8441797}, robots navigation in indoor \cite{Zhu2016TargetdrivenVN}, \cite{Chaplot2020NeuralTS} and outdoor environments \cite{sivakumar2021learned}, \cite{9064793}, \cite{9561936}. However, in practice, traditional learning approaches require access to all of the data in one place, uploaded to a central server for model training requiring high speed internet. Furthermore, robots operating in the world experience diverse and varied environments requiring continuous upload of large amounts of data to the central server. While effective in controlled environments with high bandwidths, uploading big chunks of data can be a challenge for robots in environments where high-speed internet is not available, is intermittent, outright denied or even leads to significant battery power consumption. 

Federated learning (FL) \cite{McMahan2016CommunicationEfficientLO} reduces the bandwidth requirement while enabling these robots to collectively enhance their learning by sharing model updates. With recent advancements in the capabilities of edge devices, federated learning takes advantage of edge computation to train models locally and shares model parameters instead of raw data with the server to learn a shared global model.  FL also allows robots to send updates at intervals, rather than continuously streaming data reducing bandwidth usage. Further, through federated learning, there is hope that robots can gather insights from their respective environments, while also contributing to a global pool of knowledge to learn adaptable models for varied environments on the go. Traditionally, FL learns a single model that tries to minimize the average loss across robots. However, local data on deployed robots is highly non-IID due to different usage and operating locations. During FL, the divergence of the local datasets due to their non-IID nature leads to slower convergence and worsening learning performance when the models are aggregated. In such cases, a singular global model suffers and may perform worse than local models for some robots. With non-IID data, it is improbable that there exists a single global model that fits the needs for all robots.  The global model can be biased and unfair. Current robotic systems that use federated learning frameworks do so in simulation \cite{9013081} or in structured indoor environments \cite{10025836} and do not account for heterogeneity that arises in the real-world deployment of robots. 

One way to avoid biased global models is to learn personalized models by clustering robots with similar local data distributions and training one aggregate model for each cluster. As a result, robots collaborate with only robots with similar experiences avoiding biases and negative performance. Previous clustered FL methods compare local model weights or gradients that rely on indirect information of the data distribution. \cite{sattler2020clustered} and \cite{ghosh2020efficient} cluster the clients and learn individual cluster models but incur a high communication cost in doing so. 

In this paper, we highlight the first clustering-based system, Federated-EmbedCluster(Fed-EC) for self-supervised visually guided autonomous navigation which overcomes the need for high bandwidth speeds. Fed-EC is deployed on two different visual navigation models to showcase its modularity. To overcome the negative affect of non-IID data on model performance, Fed-EC groups the aggregation of local models by looking at similarity between the local datasets. Within each cluster group, the data is similar and mimics an IID set up ensuring that model aggregation does not degrade performance. Unlike previous methods where multiple rounds are needed \cite{sattler2020clustered} or multiple models are communicated \cite{ghosh2020efficient}, in each communication round the mean embedding vector which does not incur any additional communication cost is shared along with the local model. Fed-EC does not know the cluster identities beforehand and hence simultaneously identifies clusters within participating robots and learns individual cluster models in the federated setting.

\begin{figure*}[!htbp]
   \centering
   \vspace{4pt}
    \includegraphics[width=\linewidth]{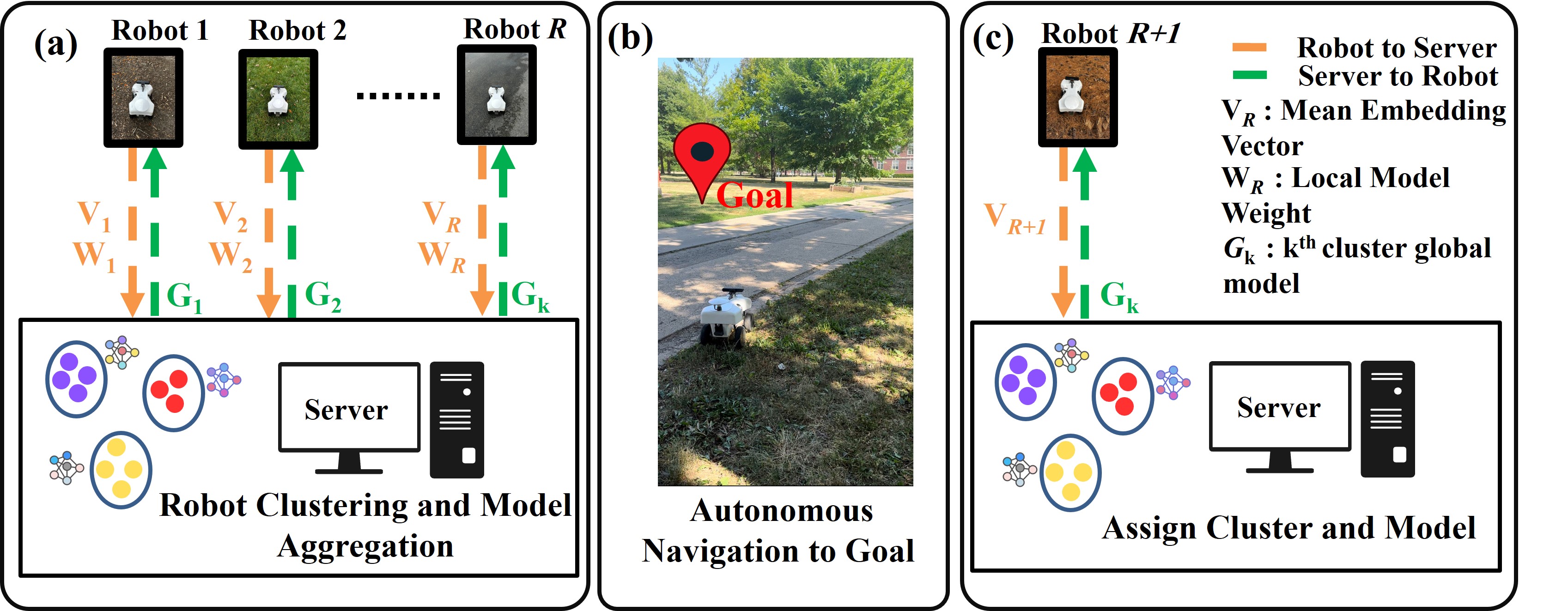}
    \caption{Workflow of Fed-EC. (a) Participating robots learn and communicates a mean embedding vector and local model weight to the server. The server clusters the robots using the mean embedding vector and aggregates local models in each cluster to learn a model which is shared with the robots based on their cluster identity. (b) The robots navigate to the a given GPS goal using the learnt model which takes as input RGB and depth images from the front facing camera. (c) If a new robot is deployed it computes a mean embedding and shares it with the server. The server assigns a cluster to the robot and sends the respective cluster model to the robot to use.}
    \label{fig:pipeline}
\end{figure*}

In this paper, we consider robots in the wild that are constantly deployed with limited hardware on board, limited communication bandwidth, and battery power. The main contributions of our papers are as follows:

\begin{itemize}
    \item We propose a clustering-based personalized FL strategy Fed-EC, to overcome the problems generated by the heterogeneous nature of robotic operations. 
    \item We implement and test the framework of Federated learning in the robotics settings, in particular on real robots using two different navigation models to navigate to a given GPS point. 
    \item  We validate through real-world robot experiments in diverse outdoor terrains that Fed-EC can perform as well as the centralized framework while reducing communication size and is better than just local training. We also show that learning a personalized FL model for each cluster is better than learning a singular global FL model over all robots. 
    \item We also show the transferability properties of our system to new robots that join the network. 
\end{itemize}

\label{sec:intro}

\section{Related Work}
\textbf{Federated Learning} Federated Learning (FL) \cite{McMahan2016CommunicationEfficientLO,Konecn2016FederatedOD,Hard2018FederatedLF,bonawitz2019towards,li2020federated}
as mentioned in \ref{sec:intro} is a distributed learning framework addressing privacy concerns, communication efficiency and enabling collaborative model training. FedAvg  introduced by \cite{McMahan2016CommunicationEfficientLO} is the most widely used FL algorithm that learns one global model for all participating clients by aggregating local model updates from them. A key problem in FL is that individual local data on clients are usually non-IID. Several research has focused on providing possible solutions for data heterogeneity in FL \cite{mohri2019agnostic,li2019convergence,8889996,li2020_Federated}. These methods learn a single global model for non-IID data. One such algorithm, proposed by \cite{li2020_Federated} bounds the differences between the local and global optimizations by introducing a proximal term to the local objective function. \cite{zhao2018federated} investigates the effect of non-IID data on FL and the degradation of performance.

\textbf{Clustered Federated Learning} Other research focuses rather on learning personalized models for each client. Clustered Federated learning (CFL) instead of optimizing a shared global model, partitions the clients into clusters and divides the optimization goal into several sub-objectives. CFL learns multiple models for each cluster,  which are more specialized and achieve better accuracy. \cite{sattler2020clustered} proposed a \textbf{CFL} algorithm that bi-partitions the clients based on the cosine similarity of their gradients and checks for congruent partition using the gradient norm. Multiple rounds are required to cluster incongruent clients which has high computational and communication costs. \cite{ghosh2020efficient} proposed Iterative Federated Clustering Algorithm \textbf{IFCA}, that initializes $k$ global models at the server which is sent to each client to compute loss. Clients are assigned to the cluster that produces the smallest loss on the local client data. This method incurs a high communication overhead due to $k$ models being broadcast. \cite{briggs2020federated} uses the difference between the local and the global model weights to perform hierarchical clustering. \cite{duan2020fedgroup} introduced \textbf{FedGroup} which uses Euclidean distance of decomposed cosine similarity metric to cluster the clients. \textbf{k-FED} designed by \cite{dennis2021heterogeneity} uses Llyod's method for K-means clustering to cluster the clients. However, the method is sensitive to the initialization of the centers and can take superpolynomial time to converge. In all previous works, the authors use the local model weights to indirectly approximate the local datasets. Moreover the previous works are not tested for robotics applications and suffer from high communication costs which is of high importance for deploying robots in the wild. 

\textbf{Federated Learning in Robotics} FL is increasingly used in numerous robotics applications. However, none of the FL applications in robotics explicitly account for non-IID data, and deployment in real-world outdoor environments which leads to changing datasets and setting up server-robot communication. \cite{8772088} uses federated learning to achieve lifelong federated reinforcement learning to improve the efficiency of robotic navigation. But the method is limited to testing in simulation and using the best model learnt in simulation to test navigation using Turtle-bot in an indoor space. Similarly, \cite{10025836} also uses federated learning for reinforcement learning for robotic swarm navigation and tests in simulation and a turtle-bot in a limited indoor real-world scenario. The method accounts for communication in simulation using communication volume but does not establish a server-robot communication in real time deployment. \cite{9457207} and \cite{nguyen2022deep} apply vanilla federated learning to autonomous driving but only test it on simulated scenarios. \cite{nguyen2022deep} uses a server-less federated learning architecture by using peer-to-peer communication instead. \cite{9560791} also uses a server-less federated architecture and instead uses a gossip-based shared data structure for trajectory forecasting. Previous methods do not personalize the federated models for heterogeneous data. Our method takes the problems of real-world federated deployment into account and learns a personal cluster model to deal with heterogeneity. 

\section{System Design}
Figure \ref{fig:pipeline}, shows an overview of Fed-EC. Data collected on each robot using an onboard RGBD camera and IMU are used to calculate the mean embedding as well as train a local traversability model. Fed-EC communicates the the local model weights and the mean embedding to the server over WiFi. The mean embeddings are used to cluster the robots and a global cluster model is aggregated for each cluster. Fed-EC sends the  respective cluster model to each robot depending on its cluster identity and training continues until the local models converge on their test data. Once a good model is learnt, given a GPS goal point, the robot navigates to the goal by leveraging traversability predictions from the local models along with the robot's kinodynamic model to solve a non-linear model predictive controller. When a new robot joins the network, Fed-EC calculates a local mean embedding and communicates it to the server to identify the robot's cluster identity and the relevant cluster model is sent to the robot.

\begin{figure}[!htbp]
  \centering
  \includegraphics[width=\linewidth]{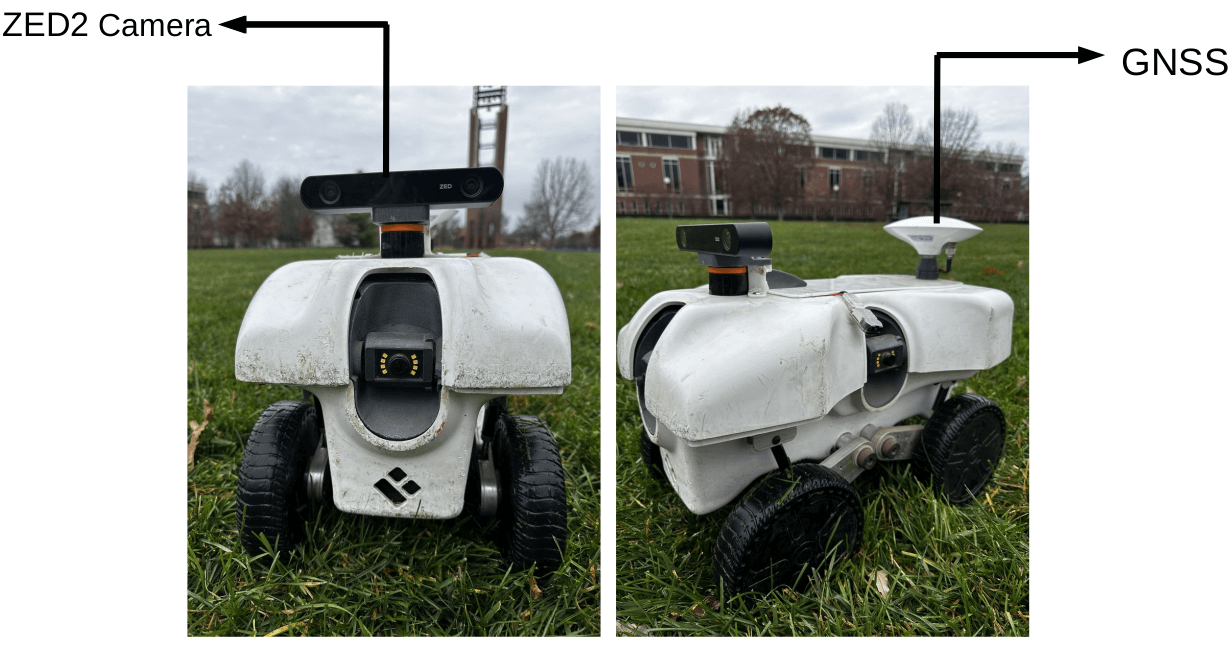}
  \vspace{-\topsep}
  \caption{Terrasentia Robot with ZED2 camera and GNSS used as a testbed for our experiments. }
  \label{fig:robot}
\end{figure}

\subsection{Robot Platform}
\label{sec:robot_platform}
The Terrasentia robot as shown in figure \ref{fig:robot}, is a small lightweight skid-steer mobile robot developed by EarthSense Inc as an agricultural robotic platform. We integrated a Jetson AGX Orin computer to run model training, store data and execute the navigation module. Terrasentia comes equipped with a 6 DOF IMU, Global Navigation Satellite System (GNSS) and wheel encoders. On top of this, we installed a front-facing ZED2 stereo-inertial camera that captures colour and depth images with a field of view of 120 degrees. The robot is also equipped with an onboard WiFi router that is used for federated learning. 

\begin{figure}[!htbp]
  \centering
  \includegraphics[width=\linewidth]{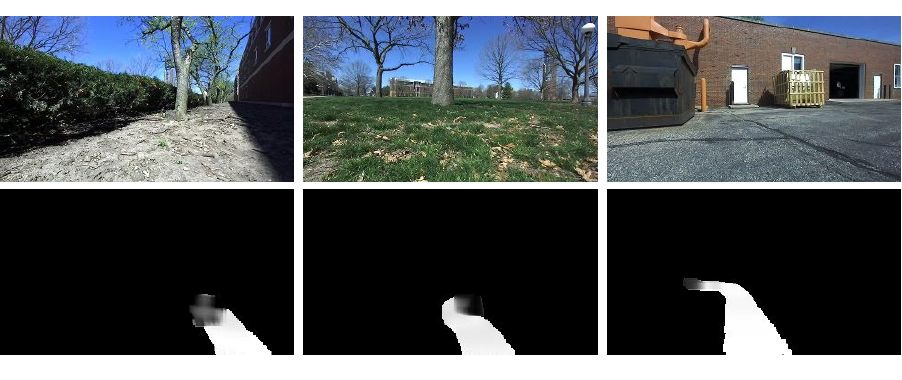}
  \vspace{-\topsep}
  \caption{Sample Data and Traversability Labels collected across different terrains. }
  \label{fig:data}
\end{figure}

\subsection{Data Collection}
\begin{figure}[!htbp]
  \centering
  \vspace{5pt}
  \includegraphics[width=\linewidth]{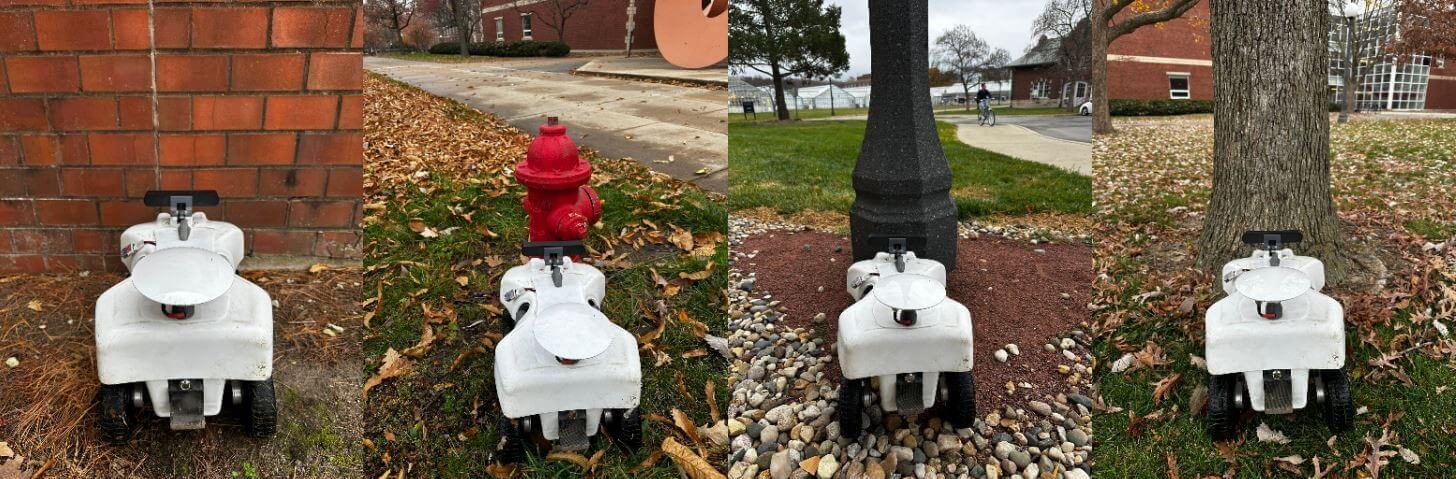}
  \vspace{-\topsep}
  \caption{Common obstacles encountered by the robot. Left to right: wall, fire hydrant,light pole, tree.}
  \label{fig:obstacles}
\end{figure}

We deploy the method described in \cite{Gasparino2022WayFASTNW} to collect data. For initial training, we manually drive the robots around in different terrains to collect input RGBD images and state estimations. This dataset includes traversability and collision labels. The traversability label is created only for the path traversed by the robot and is set to 1 for traversable and 0 for untraversable areas. Figure \ref{fig:data} shows sample data collected in three different terrains along with their traversability labels.  During data collection, we intentionally drive the robot into untraversable areas and obstacles as shown in figure \ref{fig:obstacles} to produce labels for failure cases. 

\subsection{Federated Learning with Embedding Clustering}

In this section, we present our method Fed-EC, a FL framework that address non-IID local data by clustering clients into distinct clusters based on their data distribution distances. We consider a federated learning system that consists of  $R$ robots deployed in the real world. Each robot $r$ in $R$ has access to local data $D_r$. Unlike vanilla federated learning where a unified global model is learnt, in Fed-EC the robots are grouped into k clusters at the server, and individual cluster models are learnt for each cluster.   

At round $N$ of Fed-EC, $R$ robots participate in local model training. At the $N^{th}$ round ,robot $r$ partakes in model training for $E$ epochs on its local data $D_r$ and updates local model $W_r$. Simultaneously, given an image $I_i$ in $D_r$ it is encoded into an embedding $v_i = f(I_i)$ where $f$ is a pretrained model. The robot is equipped with adequate computational power that using a pretrained model to generate embeddings is reasonable and not computationally expensive.  The embeddings are averaged to get the mean embedding  $V_r = \sum_{i=1}^{|D_r|}v_i$.  Using embeddings Fed-EC directly encodes the visual information of the local datasets. Similar embeddings represent robots deployed in similar regions or terrains. The mean embeddings are a single vector with small data sizes which are easy to upload. Each robot $r$ uploads the updated local model $W_r$  and the mean embedding $V_r$ to the server. The server deploys Density-Based Spatial Clustering of Applications with Noise (DBSCAN) \cite{Ester1996ADA} to cluster the embeddings. We deploy DBSCAN as it is able to determine the number of clusters and doesn't need to be manually defined beforehand. DBSCAN uses threshold distance between robots $r$ and $j$ as $\left \| V_r-V_j \right \|$ and minimum points in a cluster to identify regions of high density in the dataset which are separated from each other by low density regions. After determining the robot cluster identities, the local model weights within a cluster are aggregated to produce the $k^{th}$ cluster model $G_k$. The aggregation rule is simply averaging as the robots in the cluster have similar data and mimic an IID setting. The server sends respective cluster models to the robots. The model embeddings are shared each round with the server to account for changing environments and datasets. This allows the robots to be clustered in the best group for their needs. In case a robot is deployed in a vastly dissimilar area and is identified as a noise point instead of being part of a cluster, the robot uses its local model to get the best performance. Algorithm \ref{algo1} and Algorithm \ref{algo2} show the process at robot and the server.

\begin{algorithm}
\caption{Robot}
\begin{algorithmic}
\For{each epoch $e$ in $E$} 
        \State Train model over $D_r$
        \State Update local model $W_r$ 
\EndFor
\State Initialize $v_i = 0$
\For{each image $I_i$ in $D_r$}
 \State $v_i = v_i +f(I_i)$
\EndFor
\State $V_r = \frac{v_i}{|D_r|}$
\State Upload $W_r$ and $V_r$ to server

\end{algorithmic}
\label{algo1}
\end{algorithm}

\begin{algorithm}
\caption{Server}
\begin{algorithmic}
\For{$n = 1$ to $N$}
    \For{each robot $r$ in $R$ \textbf{in parallel}} 
        \State Download $W_r$ and $V_r$
    \EndFor
   \State $k$ = DBSCAN($V_1.....V_r$)
   \For{each robot $r$ in cluster $k$}
       \State $G_k = Avg(W_r)$ $\forall r \in k$
       \State Send $G_k$ to robot $r$ in cluster $k$ 
    \EndFor
\EndFor
\end{algorithmic}
\label{algo2}
\end{algorithm}

\section{Implementation}
We implemented Fed-EC on a real world robot deployed in semi-urban, forest like and urban environments. The robot as mentioned in Section \ref{sec:robot_platform} is used to collect data in different environments.The test bed consists of 1 real world robot and the collected data shuffled and then divided into 9 partitions to mimic 9 robots in different environments to study non IID behavior. The local model is trained on the edge and communicated with the server using WiFi module installed on the robot. The edge device has NVIDIA Jetson AGX Orin equipped with 12-core Arm CPU, 2048-core NVIDIA GPU and 64GB memory. It runs Ubuntu 18.04 and uses Python 3 and Pytorch for model training and inference. The server is equipped with a Intel Core i7-11800 CPU, NVIDIA GeForce RTX 3060 GPU and runs Ubuntu 18.04. 

\textbf{Local Models:} We deploy Fed-EC with two different vision navigation models: \textbf{1. WayFAST \cite{Gasparino2022WayFASTNW}},  is a convolutional neural network that takes as input RGB and depth images and outputs a traversability prediction map. The traversability prediction represents a map in the same space as the input image, that quantifies how well the robot can navigate in a given scenario. \textbf{2. BADGR \cite{Kahn2020BADGRAA}}, is an image based, action-conditioned predictive deep neural network which uses RGB images and sequence of linear and angular velocity commands to predict future collision events. 

During deployment, a model predictive controller queries the traversability values or collision values depending on the underlying navigation model and uses them to minimize a cost function that drives the robot toward a goal point while jointly maximizing traversability. As a result, areas with low traversability or collision such as trees and buildings are avoided.

\textbf{Baselines:} We compare our Federated Learning framework against three baseline frameworks: \textbf{1. Central Learning}, all of the data collected at the edge on individual robots is sent to the central server for model training, \textbf{2. Local training}, model trained on the edge on local data without collaboration between robots. \textbf{3. FedAvg \cite{McMahan2016CommunicationEfficientLO}}, the server aggregates all local models to learn one global model. 

\textbf{Training Parameters:} In all experiments, the baselines and Fed-EC use the same model architecture. The local and central baselines are trained for 30 epochs with a batch size of 16. FedAvg and Fed-EC are trained locally for 5 epochs with batch size 16. The size of the local training dataset varies between 450-1000 samples from robot to robot. The minimum distance threshold for DBSCAN is manually tuned and the minimum points in each cluster is set to 2. A pretrained ResNet-18 is used to generate the data embeddings.

\section{Evaluation}
\begin{table}[!htbp]
  \centering
  \caption{Validation Loss and IPR of different models for Traversability Prediction.}
  \begin{tabular}{ccccc}
    \toprule
    Model & Training Mode & Avg. Loss & IPR (\%) & Local Epochs \\
    \midrule
    & Local & 0.006853 & -  & 30 \\
   & Central & 0.004021 & - & 30 \\
    WayFAST & FedAvg & 0.007871 & 60 & 5 \\
    & \textbf{Fed-EC (Ours)} & 0.004621 &100 & 5\\
    \bottomrule
   & Local & 0.934291 & -  & 30 \\
   & Central & 0.226426 & - & 30 \\
    BADGR& FedAvg &  1.877938 & 40 & 5 \\
    & \textbf{Fed-EC (Ours)} & 0.530927&70 & 5\\
    \bottomrule
  \end{tabular}
  \label{tab:loss}
\end{table}

\subsection{Model Performance}

To evaluate the model performance of our method against the baselines we use validation loss averaged over all the robots and incentivized participation rate (IPR) \cite{Cho2022ToFO} which is defined as 
\begin{equation}
    \centering
    IPR = \frac{1}{M}\sum_{r=1}^{R}\mathds{1}({f_{r}^{FL}(w)>f_{r}^{local}(w)})
\end{equation}
where $f_{r}^{FL}(w)$ is the model performance on robot R using an FL strategy i.e. FedAvg or Fed-EC, $f_{r}^{local}(w)$ is the local model performance on robot R and $\mathds{1}$ is the indicator function. IPR indicates how many robots are incentivized to participate in federated learning. 

Table \ref{tab:loss} shows the validation loss on the test data across all configurations.  We perform as many epochs for local and central models as communication rounds for FedAvg and Fed-EC. The central model is trained at the server whereas all the other models are trained on the edge of the robot. The simple local baseline performs poorly and does not learn even after several epochs due to its limited data. The central baseline has access to all of the data from all the robots in one place and hence gives the ideal performance. The FedAvg model learns a single global model collaboratively, however the average loss over all the robots is higher than the local model. This is because FedAvg performs worse than the local model for robots with smaller amounts of data. Fed-EC on the other hand, clusters the robots based on their mean data embedding and learns individual personalized models for each cluster. In doing so, it reduces the loss over FedAvg and local models. Further, using Fed-EC increases the incentive for robots to choose federated learning and improves the performance by diminishing the negative transfer due to heterogeneity in data across robots. In addition, Fed-EC takes 9mins18sec for local training and computing mean embedding vector  as the data is distributed across robots which is faster than centralized method which takes 1hr12min52sec and is slightly slower than FedAvg's 8min38sec because of the embedding vector computation.

\begin{table}[htbp!]
  \centering
  \caption{Time Taken to Communicate Data From the Robot to the Server for Centralized vs Federated Learning.}
  \begin{tabular}{cccc}
    \toprule
    Learning Type & Data Type & Data Size &  Upload Time \\
    \midrule
    Central & Rosbag & 1.8GB & 00:20:08  \\
    \midrule
    Federated & \makecell{Model Weights\\ Mean Embedding Vector} & 78MB & 00:00:42\\
    \bottomrule
  \end{tabular}
  \label{tab:communication}
\end{table} 

\begin{figure}[ht!]
  \centering
  \includegraphics[width=\linewidth]{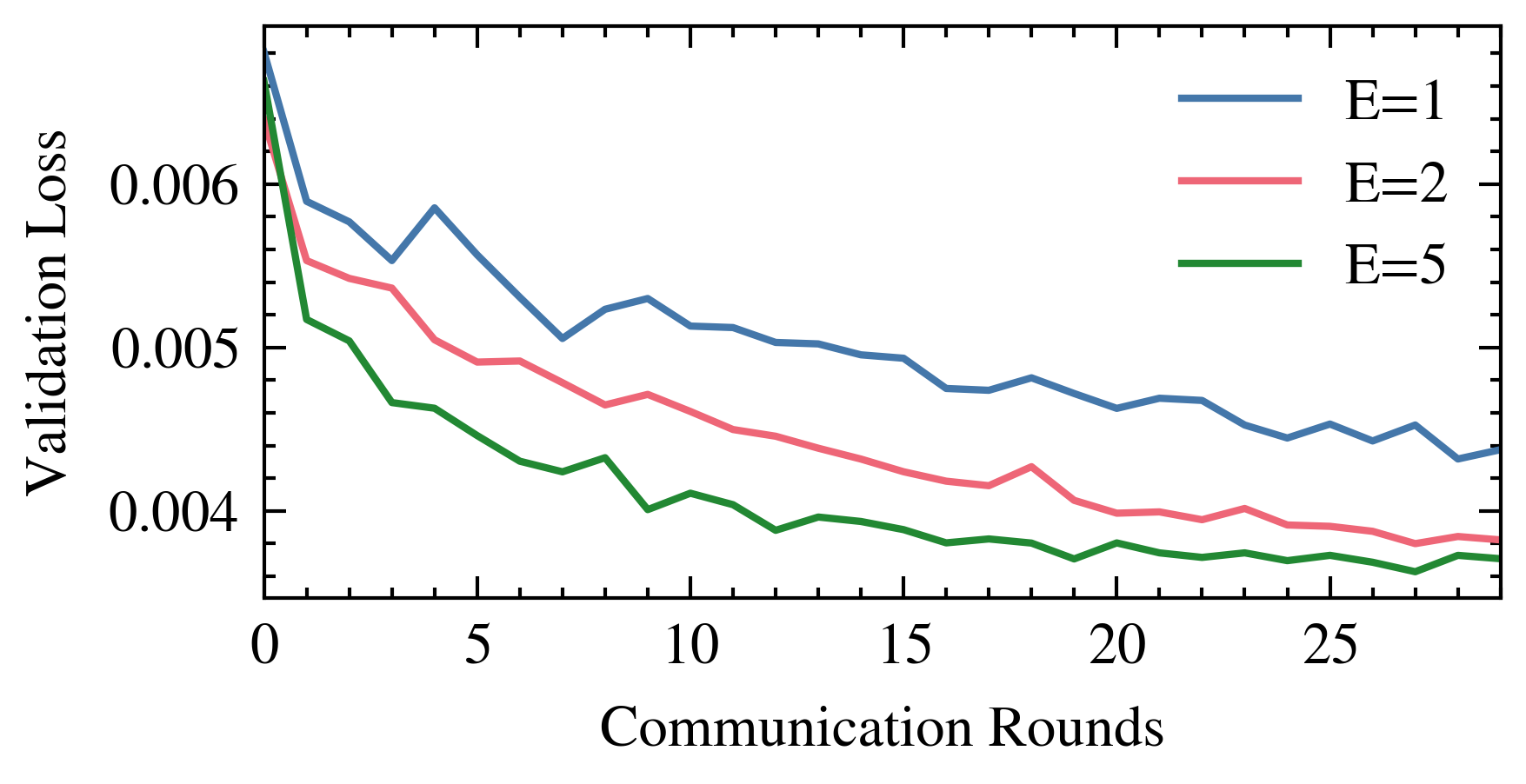}
  \caption{Effect of increasing computation per robot.}
  \label{fig:epochs}
\end{figure}

\begin{figure*}[htbp!]
  \centering
  \subfloat[]{\includegraphics[trim=0 0 0 0,clip,width=0.7\linewidth]{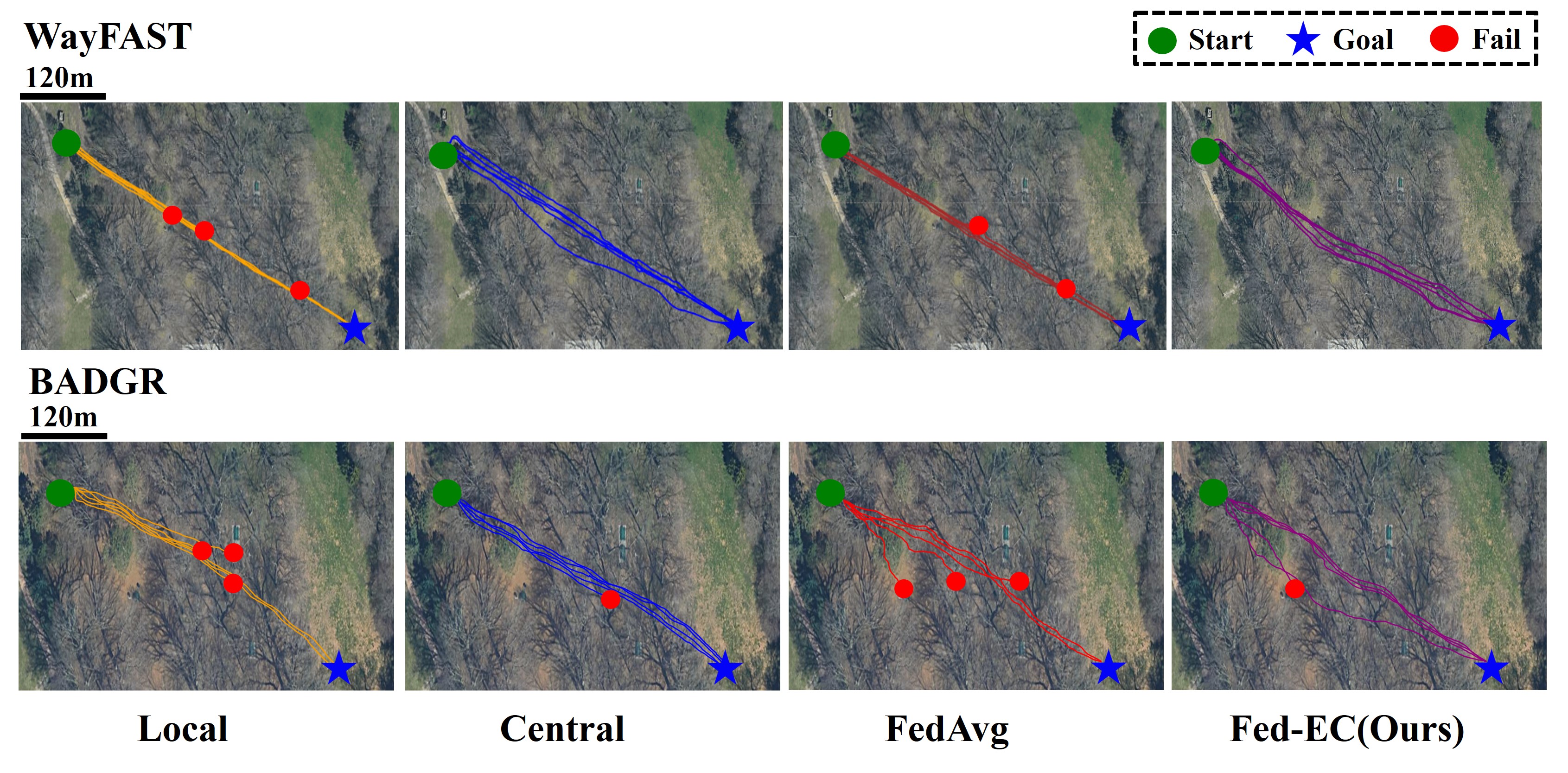}} \\
  \subfloat[]{\includegraphics[trim=0 0cm 0 1cm,clip, width=0.7\linewidth]{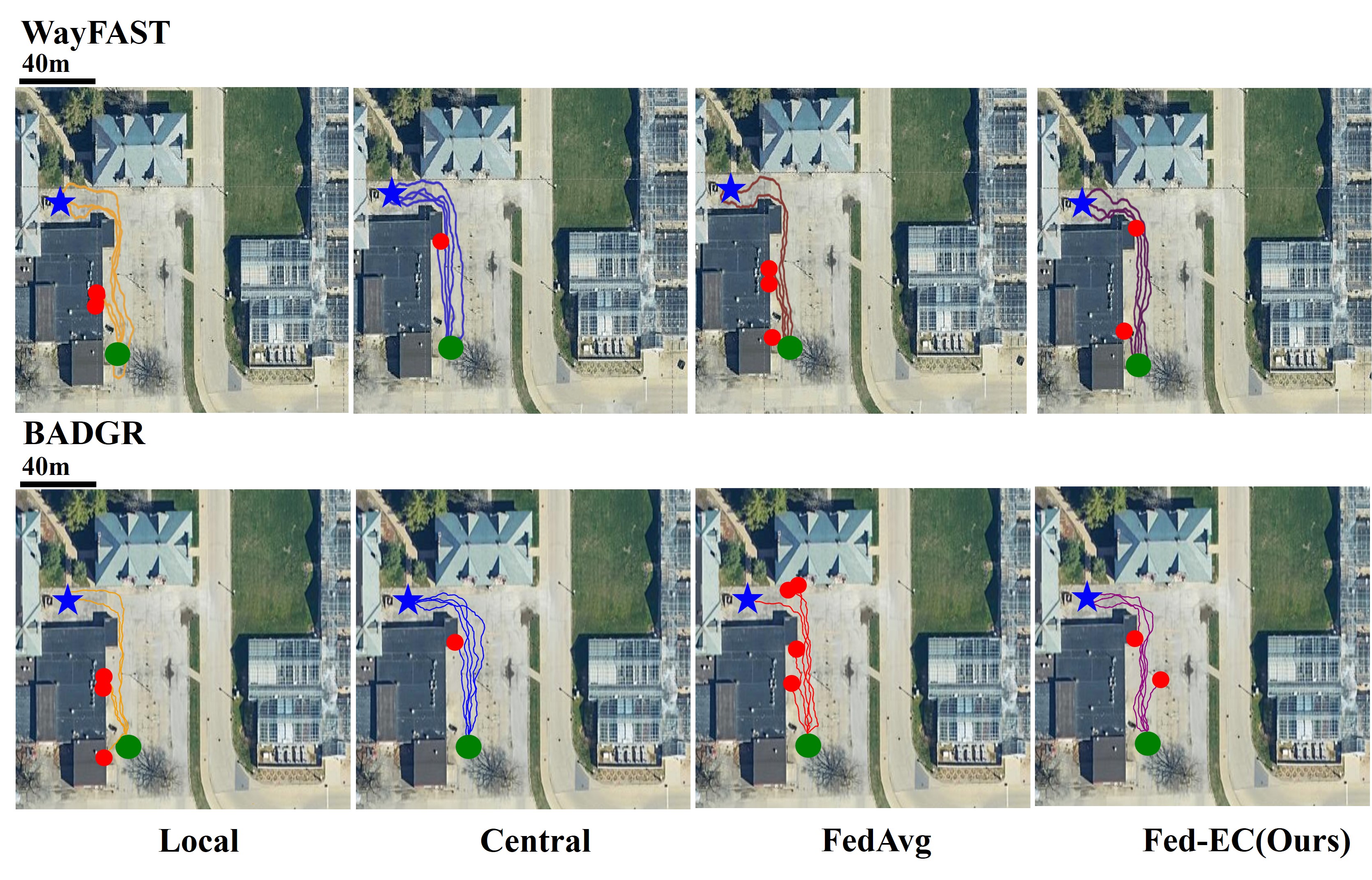}} \\
  \caption{Paths taken by Fed-EC and the baselines (Local, Central, FedAvg) to reach the goal in two different environments: (a) Forest-like and (b) Parking Lot}
  \label{fig:trials}
\end{figure*}

\subsection{Communication Cost}
The implementation of the Federated framework requires robots to communicate their model updates to the server every few rounnds. Communication efficiency is essential for training. However, communicating over the internet can cause hold-ups and delays due to the asymmetric nature of internet speeds. In general, the upload speeds are much lower than the download speeds. As seen in Table \ref{tab:communication} outdoors with limited bandwidth of 12Mbps uploading raw data to the server per robot is 23x bigger than uploading model weights. This leads to centralized learning suffering delays and bottlenecks as all robots try to upload raw data. On the other hand, model weights are uploaded in 42 seconds and due to the nature of FL can be done asynchronously while still reducing the time and need for higher bandwidth. Further, as shown in figure \ref{fig:epochs} using more computation on the edge decreases the number of communication rounds needed to reach a target loss. We increase computation on the robot by increasing the number of epochs(E) while keeping the batch size(B) constant.

\subsection{Navigation}
\begin{table}[htbp!]
  \centering
   \caption{Summary of Navigation Runs (successful runs/total runs) in three different environments. Trial 1: Forest like and Trial 2: Parking Lot.}
  \begin{tabular}{cccc}
    \toprule
    Trial&Training Mode & WayFAST & BADGR  \\
    \midrule
    &Local & 2/5 & 2/5 \\
    &Central & 5/5 & 4/5 \\
    Trial 1&FedAvg & 3/5 & 2/5\\
    &\textbf{Fed-EC (Ours)}& 5/5& 4/5 \\
    \bottomrule
    &Local & 3/5 & 2/5 \\
    &Central & 4/5 & 4/5 \\
    Trial 2&FedAvg & 2/5 & 1/5\\
    &\textbf{Fed-EC (Ours)}& 3/5& 3/5 \\
    \bottomrule
  \end{tabular}
  \label{tab:trials}
\end{table}

Table \ref{tab:trials} and figure \ref{fig:trials} show results in a forest and urban environment using two models WayFAST\cite{Gasparino2022WayFASTNW} and BADGR \cite{Kahn2020BADGRAA}. 

In the initial runs federated learning uses data previously collected through manual driving. In subsequent runs including in trial 2 the training incorporates data collected during previous runs. During each run, the robot uses the newly collected data to perform training and communicates with the server to update cluster models.  As the training and communication take time, the updated cluster models are not used in the same run but in subsequent runs of the trials. This ensures continuous, incremental improvement with each successive trial.

The local baseline performs poorly in both trials because of the limited training data on the robot. The central baseline is able to reach the goal most cases for both models as it has access to all of the data. FedAvg is able to reach the goal in trial 1, however, it fails and performs worse than the local baseline in the parking lot. This is due to the degraded performance of the FedAvg model in this environment due to the availability of limited data in this environment. Fed-EC performs similarly to the central model and is able to reach the goal in most runs without any intervention. It outperforms the local model and the FedAvg model. All the models suffer in the parking lot due to limited training data in that environment. Due to the nature of federated learning the data collected during these runs can be further used to update the cluster models (Fed-EC) and global model (FedAvg) to improve performance subsequently. 

\begin{figure}[!htbp]
  \centering
  \includegraphics[width=\linewidth]{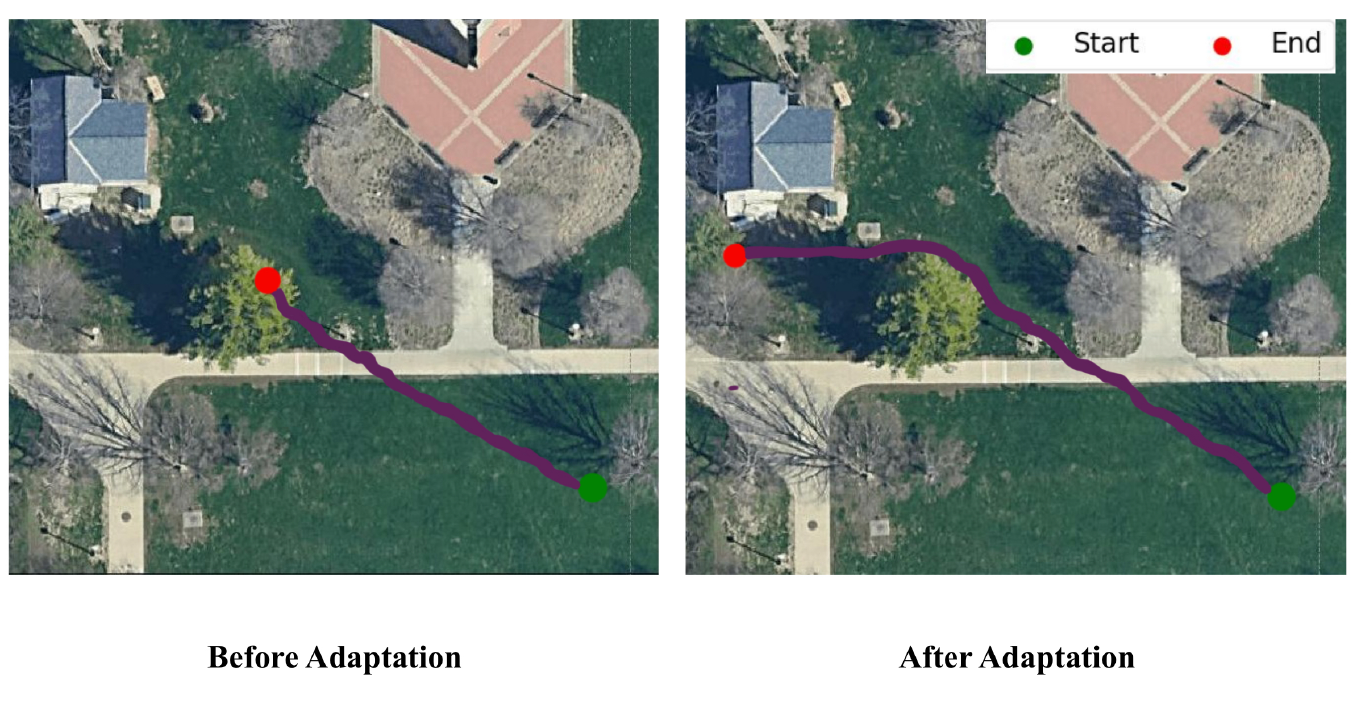}
  \caption{Robot adapts and navigates around untraversable areas.}
  \label{fig:adapt}
\end{figure}

\subsection{Adaptation}
Due to the nature of federated learning, Fed-EC is continually updating and updating the cluster model when new data is available. The learning is shared across robots in the cluster ensuring that other robots do not fail in the same scenario. Figure \ref{fig:adapt} shows an example scenario where Fed-EC can overcome obstacles by using updated models to adapt. In this case, the robot initially runs into a low-hanging tree branch, assuming it can pass below it. The robot fails, however, it records this data while navigating which is then used to train the local model and in turn update the cluster model at the server. In a subsequent run. the robot receives the new cluster model from the server and is able to recognize the branch as untraversable and moves away from it to take a different path.

\begin{figure}[!htbp]
  \centering
  \includegraphics[trim={0 6cm 0 0},clip,width=\linewidth]{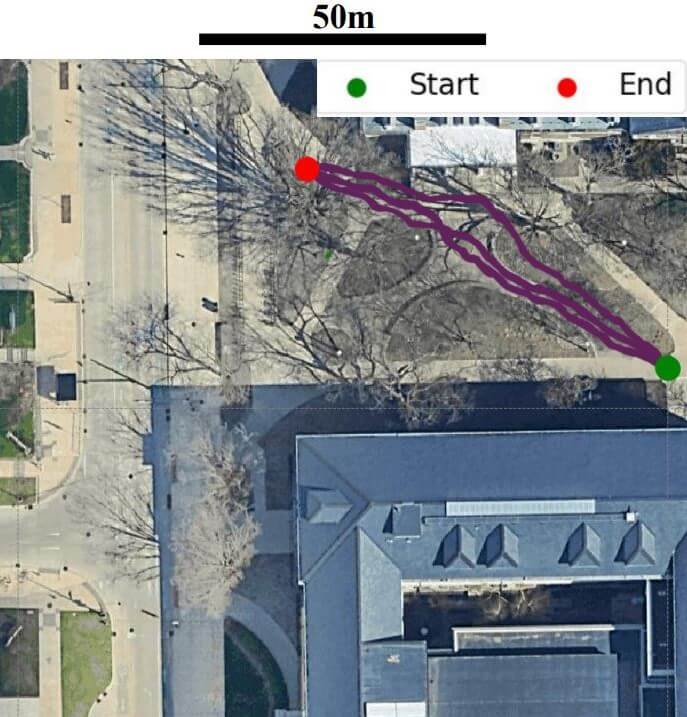}
  \caption{New robot navigates to goal using previously learnt global cluster model.}
  \label{fig:transfer}
\end{figure}

\subsection{Transferability to New Robot}
In the real world, new robots can be acquired at any point and deployed in different locations. Usually, this requires training a new model for the robot data or sending the data to a server to finetune a base model. Models like FedAvg can provide an initial model that can be finetuned on the local data. However, if the initial model is bad it will not offer significant benefit to just local training. When a new robot joins, Fed-EC can easily use the mean embedding vector of the new robot to identify which cluster it belongs to and share the respective cluster model. Fed-EC can transfer a model which can be readily used without the need for retraining. The new robot can participate in federated learning as well and better the  cluster model with its new data.  Figure \ref{fig:transfer} shows a new robot navigating using model WayFAST \cite{Gasparino2022WayFASTNW} to a goal without any intervention using the cluster model sent by the server. 

\begin{figure}[!htbp]
  \centering
  \includegraphics[width=\linewidth]{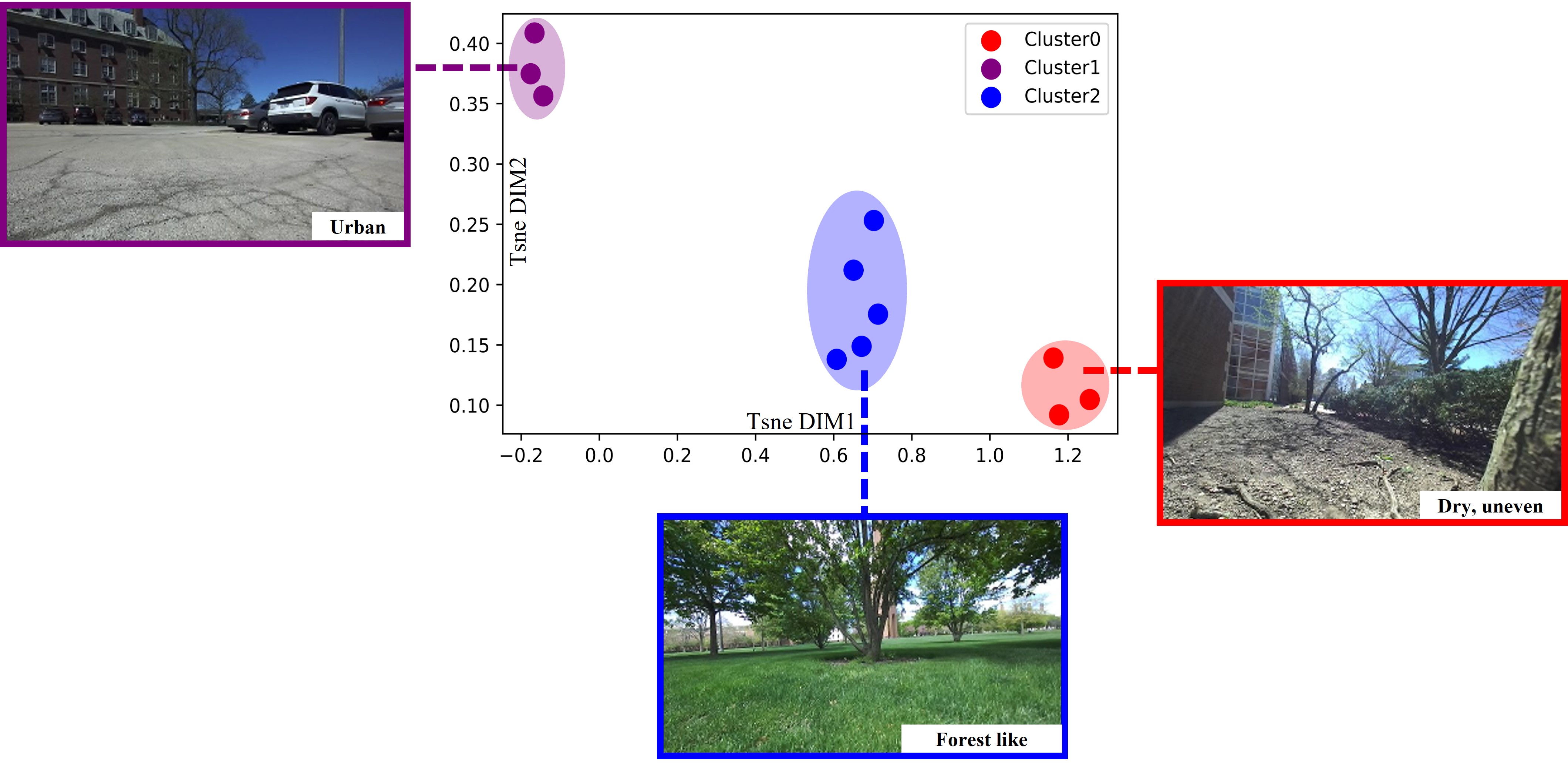}
  \caption{Visualization of 11 robots and their corresponding cluster identities using DBSCAN based on distance proximity of mean embedding vectors.}
  \label{fig:cluster}
\end{figure}

Figure \ref{fig:cluster} shows the clusters identified by Fed-EC to group the robots. We observe that robots deployed in similar terrains are grouped together as the mean embedding vectors of the local data are closer to each other while different terrains occupy separate regions on the plot. The newly deployed robot is clustered in cluster 2 in the figure and is allocated the cluster model of this group to perform navigation. 

\section{Conclusion} 
To conclude, we presented a federated learning framework that can be deployed with various navigation models on real-world robots to navigate to a given GPS point under limited communication constraints. We demonstrate that our method performs comparatively to centralized learning without the need for data to be uploaded to a server and reducing the communication size. Fed-EC also performs better than local models which are limited by no collaboration. Our framework also tackles data imbalance and outperforms FedAvg by clustering robots based on similar data distributions and learning individual personalized cluster models resulting in an unbiased global model. Further, Fed-EC continually collects data and updates the cluster models which makes it capable of adapting. Finally, when a new robot joins the network, Fed-EC clusters the robot and assigns it an initial model to use without the need for retraining. 



\bibliographystyle{IEEEtran}
\bibliography{references}

\end{document}